# Robot Mindreading and the Problem of Trust

Andrés Páez[1]

**Abstract.** This paper raises three questions regarding the attribution of beliefs, desires, and intentions to robots. The first one is whether humans in fact engage in robot mindreading. If they do, this raises a second question: does robot mindreading foster trust towards robots? Both of these questions are empirical, and I show that the available evidence is insufficient to answer them. Now, if we assume that the answer to both questions is affirmative, a third and more important question arises: should developers and engineers promote robot mindreading in view of their stated goal of enhancing transparency? My worry here is that by attempting to make robots more mind-readable, they are abandoning the project of understanding automatic decision processes. Features that enhance mind-readability are prone to make the factors that determine automatic decisions even more opaque than they already are. And current strategies to eliminate opacity do not enhance mind-readability. The last part of the paper discusses different ways to analyze this apparent trade-off and suggests that a possible solution must adopt tolerable degrees of opacity that depend on pragmatic factors connected to the level of trust required for the intended uses of the robot.

## 1 INTRODUCTION

Autonomous Artificial Intelligent Systems (AISs) designed to interact socially with humans are becoming a common presence in our private and public lives. Our increased interaction with personal virtual assistants, social chatbots, autonomous vehicles and, especially, humanoid robots invites the question of how humans interpret, predict and explain their behavior and decisions. The interpretation framework adopted will have practical effects, such as facilitating human-robot interaction and cooperation, but it will also have philosophical consequences. It will determine whether the AIS's behavior is judged to be conscious and free, and therefore subject to standards of legal and moral responsibility. The mental states attributed to AISs will also establish the rights they should be granted, and shape how humans assess the level of creativity, adaptability and potential for cooperation of an AIS. And more importantly for the purpose of this paper, the favored interpretative framework has implications for the transparency and trustworthiness of AISs, two of the main concerns of software engineers committed to the EPSRC Principles of Robotics (Boden et al. 2017) and of philosophers involved in the eXplainable AI project (XAI). Whether AISs are interpreted as intentional agents or as purely mechanic devices will affect the perception of transparency and the consequent level of trust placed in them. This paper explores the relation between interpretative frameworks and their effects on our trust in AISs.

I will use the expression *robot mindreading* to designate the attitude of attributing mental states to AISs in order to explain and predict their decisions and actions.[2] According to the conventional meaning of 'mindreading' (Nichols & Stich 2003), successful interaction with others involves the attribution of beliefs, desires, emotions, and intentions to make sense of their behavior and predict their future actions. Many researchers have defended the idea that humans naturally engage in robot mindreading. For example, de Graaf and Malle argue that "systems that are in fact autonomous and intelligent will almost always exhibit some indicators of intentional agency (e.g., initiative, planning, decision making), and as soon as these indicators lead people to actually regard them as intentional agents, people will apply the human conceptual framework of behavior explanation to them" (2017, p. 19).

However, the evidence for robot mindreading is not conclusive. Critics have questioned the idea of robot mindreading because of the self-report method used to evaluate the spontaneous attribution of mental states to robots (Scholl & Tremoulet 2000). Neuroimaging studies have also thrown doubts on the idea (Chaminade et al. 2012). The first task of the paper will thus be to examine the evidence and determine in which sense and to what extent humans engage in robot mindreading. This will be the topic of section 2.

The main question I want to address, however, is whether developers and engineers *should* promote robot mindreading in view of their stated goal of enhancing transparency and trust. The question has an empirical component and a theoretical component. The empirical component depends on the results of section 2: if humans seldom engage in robot mindreading, it would be useless to promote mindreading as a means to generate trust. If they do so often, it is still a separate empirical question whether mindreading has the effect of creating trust. On the theoretical side, the question is whether transparency and mindreading-based trust are compatible goals in the case of robots. The risk is that by attempting to make AISs more mind-readable, we are abandoning the project of understanding automatic decision processes. Features that enhance mind-readability are prone to make the factors that determine automatic decisions even more opaque than they already are. And current strategies to eliminate opacity do not enhance mind-readability. The last part of the paper discusses different ways to analyze this apparent trade-off and suggests that a possible solution must adopt tolerable degrees of opacity that depend on pragmatic factors connected to the level of trust required for the intended uses of the AIS.

---

[1] Department of Philosophy, Universidad de los Andes, Colombia. Email: apaez@uniandes.edu.co

[2] In the literature on human-robot interaction it has become increasingly popular to talk about taking "the intentional stance" (Dennett 1987) towards AISs. In the next section I explain why I prefer the more theoretically neutral term *mindreading*.

## 2 THE EVIDENCE FOR ROBOT MINDREADING

Humans readily attribute mental states to other humans, to non-human animals (Mameli & Bortolotti 2006), and even to abstract shapes (Heider & Simmel 1944). It thus seems natural to assume that they also spontaneously attribute mental states to robots. This assumption has been strengthened by recent developments in robot design that aim at facilitating meaningful interaction between robots and humans. The goal for many researchers in robotics is to create multimodal interfaces that closely mimic human appearance, behavior and speech to provide social communicative functionality that is natural and intuitive (Duffy 2003). Social bots can now evaluate the emotional state of a human and adjust their behavior to build rapport and appear empathic (Novikova & Watts 2015). AISs can also rationalize their decisions by translating their internal state-action representations into natural language (Ehsan et al. 2018). And we must not forget that humans have been primed by pop culture and science fiction to regard robots as autonomous intentional agents.

Before we approach the question of whether humans in fact engage in robot mindreading, it is important to examine two questions. The first one regards the nature of mindreading itself: What assumptions about human nature and what aspects of behavior and social interaction guide the attribution of a mental state to another person? This question is essential to robot mindreading because we need to be clear about our assumptions about the nature of AISs, and about the influence of pragmatic and social factors in the attribution of mental states to robots. The second question regards the experimental evidence for human and robot mindreading. Can introspection give us a sufficient basis to understand how mindreading works? What does neuroimaging tell us about the interpretative framework adopted to interact with the subjects to which we attribute mental states?

It has become increasingly popular to talk about taking "the intentional stance" towards robots. The phrase, coined by Dennett (1987), is often left undefined in the human-robot interaction literature, so it is worth taking a closer look at its intended meaning. According to Dennett, taking the intentional stance towards $x$ requires that we attribute to $x$ the mental states that it would be *rational* for $x$ to have given $x$'s behavior, context and needs. An initial problem for this normative characterization of mindreading is that it requires a robust meaning for 'rationality', and none has been forthcoming. Dennett himself settles for a "flexible" (p. 94) and "slippery" (p. 97) notion. But as Nichols and Stich argue, even if we reached a reasonable construal of rationality, "a plausible case can be made that mindreading does *not* depend on an assumption of rationality" (1993, p. 144). Mental states attributed to a person on the basis of his behavior can be explained by attributing an irrational desire to him. For example, if Jones knows that doing $a$ is detrimental to his well-being and he does it nonetheless, we can naturally explain his behavior by attributing to him a desire to do $a$, but that desire does not fit into a pattern of rationality. A similar case occurs with belief attribution. If Jones holds an irrational belief that results from a cognitive bias, his belief can be explained by attributing to him a thin grasp of the laws of logic or probability, but the explanation requires that we attribute to him a certain degree of irrationality. Knowledge of the most common heuristics and biases can even allow us to predict the behavior of most people in solving simple tasks involving probabilities. Finally, when an observer detects that an agent is being deceitful in any way, the observer will attribute beliefs to the agent in ways that a rationality-based theory cannot explain (Eckman 1985).

A defender of taking the intentional stance towards robots, of which there are many, could argue that robots are purposefully designed to be rational and that the assumption of rationality is granted in this case. But that would only be true if current robots were designed according to logical rules and saintly purposes. Present day AISs are based on a combination of several deep learning systems that can deviate from any watered-down notion of rationality and that inherit the biases of their trainers, as has been amply documented. There are also robots capable of deception and psychological manipulation (Wagner & Arkin 2011). Although deception has evolutionary advantages for the deceiver, it would be excluded from any construal of rational behavior that includes an ethical component.

The idea of taking the "intentional stance" towards robots is attractive because it offers a catch-all approach that simplifies our understanding of human-robot interaction. But fifty years of research in social cognition have shown that the concepts and mechanisms used to explain and predict human behavior can be as diverse as the behaviors themselves. They vary as a function of a person's mindreading goals, and are affected by other aspects of social interaction, such as social categorization, stereotypes, social biases, and situational context (Spaulding, 2018). This means that it is a mistake to try to adapt a global theory of human mindreading, be it the intentional stance or any other, to robot mindreading. AISs have been designed for specific purposes, with specific human-like features, and with abilities that have been tailored for specific users. By conducting research within these boundaries, it will be much easier to determine whether humans engage in robot mindreading, instead of pursuing strategies that promote a vaguely defined intentional stance.

Let us turn now to the question of the experimental evidence for human and robot mindreading (see Pérez-Osorio & Wykowska 2019 for a complete survey). Here we find two methodological problems. In many studies of human-robot interaction, the method used to evaluate the spontaneous attribution of mental states is based on self-reports. This method has been criticized because it does not rule out the operation of other higher-order cognitive mechanisms. For example, participants might infer the attributions from the questions of the task rather than report perceived attributions (Sholl & Tremoulet 2000). A different approach is based on neuroimaging techniques that identify the neural mechanisms underlying mindreading. During the last 30 years, the physiological basis of mindreading has been detected by comparing brain imaging studies of people with autism, whose mindreading abilities seem to be faulty to varying degrees, with those of normal volunteers (Frith & Frith 2000). However, these studies are still at a very early stage. They can only imperfectly detect which areas of the brain correspond to the ability to mindread, because not all social, linguistic, and emotional behavior require this ability. More recently, researchers have tried to detect "on-line" mindreading in normal volunteers using different tasks involving goal-directed behavior, such as understanding a story that requires the attribution of desires and intentions to its characters (Fletcher et al. 1995), second-guessing an opponent in a game (Gallaher et al. 2002), and detecting stimuli that signal the actions of another individual (Allison et al. 2000). Using these methods, a correlation between mindreading and activity in several brain regions has been claimed. The problem, again, is that this claim depends on the assumption that the subjects in these studies in fact are not using other higher-order mechanisms to perform the tasks. Despite these promising results, it must be said that the use of neuroimaging to understand mindreading is still in its infancy.

Bearing in mind these methodological limitations, let us consider some of the evidence for and against robot mindreading based on neuroimaging. It has been shown that AISs elicit emotional responses and social behavior akin to those caused by other humans (Appel et al. 2012, Rosenthal-von der Pütten et al. 2014), and there is physiological evidence that humans can empathize with the perceived pain of robots (Suzuki et al. 2015). There is also evidence that the degree of anthropomorphism and embodiment of AISs is positively correlated with the activation of brain regions associated with the inference of intentions, goals and desires in others (Hegel et al. 2008). More generally, anthropomorphism relies on the same cognitive mechanisms that generate the attribution of intentions to human behavior (Castelli et al. 2000). On the other hand, Krach et al. (2008) showed that areas of the brain previously shown to have been associated with mindreading were not activated in response to artificial agents regardless of their human-like appearance. And Chaminade et al. (2012) shows that in a rock-paper-scissors game, people's brains react differently depending on whether they believe they are playing against a human or an AIS. In conclusion, the evidence thus far for and against robot mindreading is unimpressive. If the study of human mindreading is its infancy, the study of robot mindreading is a research project waiting to be executed.

## 3 ROBOT MINDREADING AND TRUST

There is no doubt that the attribution of beliefs, desires and intentions to an AIS facilitates human-robot interaction (Fong et al. 2003, Fink 2012). A gamer's experience will be enhanced if she believes that her artificial opponent has (evil) intentions and desires, and a companion robot will better achieve its purpose if its owner believes that the robot actually cares about his woes. The general idea is that the cognitive and emotional response to robots will be more positive if the user treats it as an intentional agent.

The first question I want to address is whether robot mindreading also fosters *trust*. I will assume in what follows that one of the main goals of robotics in particular, and of developers of AI in general, is to increase public trust towards artificial intelligent systems. Distrust in AISs can take different forms. One source of concern among the public is the danger posed by biased algorithms, which seem to garner much attention from the press. Governments and the private sector have taken strides to address the ethical challenges posed by AI because they are aware that public trust is essential for the consolidation of the so-called 4th Industrial Revolution. A different source of concern is the perception that decisions made by an automatic system are not reliable, even when unbiased, and should not be trusted. Patients are reluctant to use health care provided by medical artificial intelligence even when it outperforms human doctors (Longoni et al. 2019) and most people do not trust automated vehicles (Hutson 2017). In sum, there is a concern about the intended or unintended biases implanted in robots by their developers, and about their reliability and performance.

In human-human interaction, honesty, competence and value similarity are essential to establish both cognitive and emotional trust (Gambetta 1991). Cognitive trust is based on one's knowledge and evidence about the trustee, about his or her reliability, while emotional trust is based on the feelings generated by our interactions with others. Honesty, competence and value similarity can only be ascribed to others by attributing to them the adequate intentions and beliefs from which these traits can be inferred. Prima facie, then, enhancing traits that convey intentions and beliefs conducive to the creation of trust should be a goal of robotics. The initial question can now be restated in the following terms: Will people be more trustful towards an AIS if its decisions and behavior are seen as the result of mental states from which honesty, competence and value similarity can be inferred? Intuitively the answer should be affirmative. If a robot behaves in ways that resemble to a high degree those of a trustworthy human, and if the user makes sense of the AIS's behavior by attributing mental states to it, there is no reason to believe that the user will not trust the AIS.

The main problem with this answer is that it extrapolates the trust-building features of human relations to the field of robotics without having enough empirical support. A meta-analysis of factors affecting trust in human-robot interaction (Hancock et al. 2011) revealed that "robot characteristics, and in particular, performance-based factors, are the largest current influence on perceived trust in HRI" (p. 523). This finding is in line with the performance-based definitions of trust found in the literature on multi-agent systems (Witkowski et al 2001).

Hancock et al. also found that factors related to human attitudes towards robots had a small role in trust building. The authors do not conclude that human factors have no influence on HRI. "Rather, the small number of studies found in this area suggests a strong need for future experimental efforts on human-related, as well as environment-related, factors" (p. 523). It could be argued that this meta-analysis focused on cognitive trust, ignoring the fact that emotional trust is more likely to be detected as an effect of robot mindreading. There are in fact several studies about the emotional reaction of humans towards robots (Rosenthal-von der Pütten et al. 2013), and there is anecdotal evidence of emotional attachments to robots (Klamer et al. 2011). However, none of these studies have measured emotional trust as an independent variable, so it is impossible to draw any conclusions about the relationship between mindreading and emotional trust.

In sum, the empirical evidence for the trust-building effects of people's attitudes towards robots, and in particular, of the interpretative framework adopted towards them, is quite thin. I should add that in most cases humans are plainly aware that they are interacting with an artificial being that lacks intentions, consciousness, desires and free will. Despite attributing mental states to machines as an expedient means to predict and explain their behavior in certain contexts, humans are still able to identify true intentional systems. More importantly, momentary rapport and fluid interaction do not entail overall trust and understanding. Trust is not directed towards the individual decisions of an AIS but rather towards its global performance. And the sense of understanding that arises from attributing mental states to an AIS can quickly disappear when the machine behaves in unexpected ways. In sum, the answer to this question is as uncertain as the answer to the question discussed in the previous section. We do not know if humans in fact engage in robot mindreading, or the specific circumstances in which they might do so, and neither can we say with any degree of certitude whether robot mindreading will foster trust towards AISs.

## 4 ROBOT MINDREADING AND OPACITY

Now, assume for a moment that robot mindreading in fact builds trust, i.e. that there are certain design features of robots that make it easier for people to attribute to them trust-conducive mental states. Working under this assumption we can now ask if the field of robotics should work towards enhancing mindreading. The main reason for raising this question is that the problem of trust in AI systems has a flip side. According to many recent papers that advance the research agenda of XAI (Ribeiro et al. 2016, Doshi-Velez & Kim 2017, Samek et al. 2017, Gilpin et al. 2018, Guidotti et al. 2018, Edmonds et al. 2019, Páez 2019), to trust an AIS is to understand its actual decision-making process, to make it explainable, transparent, comprehensible and interpretable.[3] Transparency and trust go hand in hand. The final question I want to address is whether this second source of trust is theoretically and practically compatible with the goal of promoting the attribution of trust-conducive mental states to robots. If they are not, which one should prevail? Is it possible to develop them in complimentary fashion?

In many cases, the question of trust in AISs is not accompanied by a demand for transparency. For example, when the goal of a humanoid robot is to provide emotional support, users need to feel that their social companion is empathic and understanding. Otherwise they will stop using it. This is a form of interaction that requires trust, in particular, trust in the judgments, perceptions and advice of the AIS, but it is unlikely that users will feel the need to know how these are reached. In fact, the AIS's utility may be negatively affected by increased transparency (Wortham & Theodorou 2017). However, since many social robots are used in healthcare environments, providers and regulators will want to know whether the content that the AIS is transmitting to a patient, a child or a senior in a vulnerable emotional or physical condition promotes their emotional wellbeing and is not detrimental to their mental health. Thus, healthcare professionals will also seek transparency in the decision-making process of social robots.

Furthermore, according to the 4th Principle of Robotics crafted by EPSRC and AHRC (Boden et al. 2017),

> Robots are manufactured artefacts. They should not be designed in a deceptive way to exploit vulnerable users; instead their machine nature should be transparent. … although it is permissible and even sometimes desirable for a robot to sometimes give the impression of real intelligence, anyone who owns or interacts with a robot should be able to find out what it really is and perhaps what it was really manufactured to do (p. 127).

Is making robots more mindreadable a violation of this principle? Is mindreadability a kind of deception? The principle allows for robots that give the impression of real intelligence, but at the same time there must be a way to make their decision processes transparent. Is it possible to have it both ways?

Although they both aim at trust-building, transparency and mind-readability are goals that pull in different directions. The purpose of the addition of features that promote the attribution of beliefs and intentions to an AIS is to facilitate the kind of interaction and closeness that leads to emotional trust, and that allows the user to make sense of its decisions. But the search for an explanation for the decisions of an AIS aims at a different goal: to make sure that its decisions are *warranted*. Transparency generates what I will call *objective trust* in AISs (Witkowski & Pitt 2000, Tong et al. 2013). Chances are that in order to achieve one goal, developers will sacrifice the possibility of achieving the other.

Robot systems are still in their infancy in terms of their ability to accurately explain their own behavior, especially when confronted with noisy sensory inputs and executing complex sequential decision processes (Edmonds et al. 2019). Attempts to explain a robot's decisions and behavior using data-driven approaches are likely to fail given the noisy inputs (Anjomshoae et al. 2019), but the possibility of designing a "transparent robot" is an ongoing research project with some promising results (see below).

However, the current trend in HRI is to design robots that offer natural language explanations that do not purport to represent their inner state or describe their sequential decision processes. Instead, the idea is to offer the explanation that a human would offer when performing a similar action. This idea has been labeled "explainable agency" (Langley et al. 2017).

Consider two recent examples of this approach. Ehsan et al. (2018) introduce what they call "AI rationalization":

> AI rationalization is a process of producing an explanation for agent behavior as if a human had performed the behavior. AI rationalization is based on the observation that there are times when humans may not have full conscious access to reasons for their behavior and consequently may not give explanations that literally reveal how a decision was made. In these situations, it is more likely that humans create plausible explanations on the spot when pressed (p. 81).

There is no intention to make AI rationalizations an accurate representation of the true decision-making process. Instead, rationalization sacrifices accuracy for real-time responses, is more intuitive to non-expert humans and will generate higher degrees of satisfaction, confidence, rapport, and willingness to use autonomous systems.

Hellström & Bensch (2018) defend a similar approach in which "understanding a robot" means having a successful interaction with it. And achieving a natural, efficient and safe interaction requires mindreading:

> Understanding of a robot is not limited to physical actions and intentions, but also includes entities such as desires, knowledge and beliefs, emotions, perceptions, capabilities, and limitations of the robot. ... Hence, we say that a human understands a robot if she has sufficient knowledge of the robot's [state-of-mind] in order to successfully interact with it. (pp. 115-116).

---
[3] Each of these terms has been fleshed out in different ways in the literature. I will use 'transparency' as a catch-all term for all of these variants. See Lipton (2018) for a comprehensive analysis.

A common assumption of both accounts is that robot mindreading is useful to fulfill the intended purpose of the AIS. But usefulness is an interest-relative notion. Robot mindreading is not useful at all for a developer trying to debug or improve the reliability of a robot. Thus usefulness–or utility–is one of the keys to understanding the relation between transparency and mindreading. For some agents it is useful to tolerate a high degree of opacity; for some, it is not useful at all.

The other key to the relation is risk. If robots are perceived as intentional agents, their actions have real effects on the psyche of its users, as we saw in the case of social robots used in healthcare environments. This means that robot designers have a responsibility towards vulnerable users of the robot that goes beyond the intended goal of providing companionship and entertainment. Their responsibility is to guarantee to a reasonable degree that the actions of the robot will not be detrimental to the patients, and this can only be achieved by understanding the underlying decision processes.

Thus, both dimensions have to be considered in robot design and implementation. A robot can fulfill the utility dimension to a very high degree while obtaining a high grade on a risk scale. What should the recommended course of action be? There is no algorithm that can help us decide which dimension should prevail. It depends on the kind of utility, the kind of risk, the needs of the users and the risk aversion of the people responsible for the implementation of the robot. Therefore, the resolution of the tension between mindreading and transparency is pragmatic through and through.

Seen from another angle, the relation between mindreading and transparency has an ethical side. The rationalizations offered by a robot are, strictly speaking, a false account of its decision process, but they are offered to the user without disclaimer to make her interaction with the robot easier. In a sense, we have created lying robots. Should this disqualify them as morally worthy companions? Zerilli et al. (2019) have expressed their concern "that automated decision-making is being held to an unrealistically high standard here, possibly owing to an unrealistically high estimate of the degree of transparency attainable from human decision-makers" (p. 661). Should we then tolerate the same level of insincerity that we find in human-human interactions?

The philosophy of testimony offers a possible answer to this question. According to the anti-reductionist position about testimony, human communication would be impossible if we did not have a natural tendency to believe what other people say without demanding justification at every junction. In Tyler Burge's words, "a person is a priori entitled to accept a proposition that is presented as true and that is intelligible to him, unless there are stronger reasons not to do so" (1993, p. 469). Among the reasons to doubt a testimony are clear signs of insincerity or incompetence or both. A user interacting with a social robot could also claim a presumptive right (Fricker 1995) to believe the reasons it offers to explain its behavior, unless the reasons are perceived as obviously false or nonsensical.

But this answer is insufficient. In high-stakes situations, such as those encountered in law, finance or medicine, a user will demand that the reasons offered match the underlying decision processes. It will not be enough that the explanations offered make sense and seem true. In these areas, procedure, evidence, statutes, and precedent are necessary elements of a satisfactory explanation. In the parlance of philosophers of testimony, the testimony has to be "reduced" or justified. Robots also have to make high-stakes decisions that require complex explanations not likely to be delivered in the form of friendly chatter or ready-made explanations. The use of real-time graphical outputs to represent the internal states and decision-making processes taking place within a robot seems to be a promising road to robot transparency (Wortham et al. 2017, Edmonds et al. 2019). This approach does not require the use of mindreading-friendly features. Quite the contrary. By making explicit the robot's software hierarchical architecture, it makes it difficult to think of the robot as a being with human-like mental states.

## 5 CONCLUSIONS

Many intuitions about our interaction with robots might turn out to be right, but it is important to verify them empirically. My first goal in this paper has been to call attention to the lack of empirical evidence for human and robot mindreading and for its success as a trust-building mechanism. Even if robot mindreading turns out to be an effective way to generate subjective trust, this goal has to be balanced against other competing goals such as transparency and objective trust. There is no formula that can determine how to weigh these factors, and it is necessary to acknowledge that pragmatic factors will inevitably decide the way forward.

I do not want to claim that transparency and subjective trust are incompatible in principle. Some authors remain confident that it is possible to create transparent robots that are nevertheless emotionally engaging and useful tools across a wide range of domains (Wortham & Theodorou 2017). But it is important to recognize that mindreading and transparency are in tension and that there are practical, theoretical and philosophical obstacles that must be overcome before this tension can be resolved.